\documentclass[runningheads]{llncs}
\usepackage{graphicx}
\usepackage{amsmath,amssymb} 
\usepackage{color}
\usepackage[width=122mm,left=12mm,paperwidth=146mm,height=193mm,top=12mm,paperheight=217mm]{geometry}

\usepackage{multirow}
\usepackage{booktabs}
\newcommand{\specialcell}[2][c]{%
  \begin{tabular}[#1]{@{}c@{}}#2\end{tabular}}

\newcommand{\scal}[1]{\mathit{#1}}
\newcommand{\vect}[1]{\mathbf{#1}}
\newcommand{\matr}[1]{\mathbf{#1}}

\newcommand{\set}[1]{\mathcal{#1}}

\begin{document}
\pagestyle{headings}
\mainmatter

\title{Multi-Task Zero-Shot Action Recognition with Prioritised Data
  Augmentation} 

\titlerunning{Multi-Task Zero-Shot Action Recognition}

\authorrunning{X. Xu et al.}

\author{Xun Xu, Timothy M. Hospedales and Shaogang Gong}


\institute{School of Electronic Engineering and Computer Science,\\
	Queen Mary University of London\\
	\email{ \{xun.xu,t.hospedales,s.gong\}@qmul.ac.uk}
}

\maketitle

\begin{abstract}
Zero-Shot Learning (ZSL) promises to scale visual recognition by
bypassing the conventional model training requirement of annotated examples for every
category. This is achieved by establishing a
mapping connecting low-level features and a semantic description of
the label space, referred as visual-semantic mapping, on auxiliary data. Re-using the learned
mapping to project target videos into an embedding space thus allows novel-classes to be
recognised by nearest neighbour inference. However, existing ZSL
methods suffer from auxiliary-target domain shift intrinsically induced
by {assuming} the same mapping for the disjoint auxiliary and target
classes. This compromises 
the generalisation accuracy of ZSL recognition on the target data.  In this
work, we improve the ability of ZSL to generalise across this domain
shift in both model- and data-centric ways by formulating a visual-semantic mapping with better generalisation properties and a
dynamic data re-weighting method to prioritise auxiliary data that are
relevant to the target classes. Specifically: (1) We introduce a
multi-task visual-semantic mapping to improve generalisation
by constraining the semantic mapping
parameters to lie on a low-dimensional manifold, (2) We explore prioritised data 
augmentation by expanding the pool of auxiliary data with additional
 instances weighted by relevance to the target domain.
The proposed new model is applied to the challenging zero-shot action
recognition problem to demonstrate its advantages over existing ZSL models. 


\end{abstract}

\section{Introduction}
Action recognition has long been a central topic in computer vision
\cite{Aggarwal2011}. A major thrust in action recognition is scaling
methods to a wider and finer range of categories
\cite{Schuldt2004,Kuehne2011,Soomro2012}. The traditional approach to
dealing with a growing number of categories is to collect labeled
training examples of each new category. This is not scalable,
particularly in the case of actions, due to the temporally extended
nature of videos compared to images, making annotation (segmentation
in {\em both} space and time) more onerous than for images.  In
contrast, the Zero-Shot Learning (ZSL) \cite{Lampert2009,Socher2013}
paradigm is gaining significant interest by providing an alternative
to classic supervised learning which does not require an ever
increasing amount of annotation.  Instead of collecting training data for the
target categories\footnote{Target and testing all refer to
  categories (e.g. action classes) to be recognised without labelled
  examples.} to be recognised, a 
classifier is constructed by re-using a visual to semantic
space mapping pre-learned on a training/auxiliary set
\footnote{Auxiliary and
  training all refer to categories (e.g. action classes) with labelled
  data.} of totally independent (disjoint) 
categories. Specifically training class labels are represented in a vector space such as attribute \cite{Lampert2009,Akata2015} or word-vectors
\cite{Socher2013,FuXKG_CVPR15}. Such vector representations of class-labels are referred to as {\em semantic label embeddings} \cite{Akata2015}. A mapping (e.g. regression \cite{Xu2015} or bilinear
model \cite{Akata2015}) is learned between low-level visual features
and their semantic embeddings. This mapping is assumed to generalise and be re-used to project visual features of 
target classes into semantic embedding space and matched against target class embeddings.
 
A fundamental challenge for ZSL is that in the context of
supervised learning of the visual-semantic mapping, the ZSL setting
violates the traditional assumption of supervised learning
\cite{Pan2010} -- that training and testing data are drawn from the
same distribution. Thus its efficacy is reduced by \emph{domain shift}
\cite{Fu2015,dinu2014improving,Lazaridou2014}. For example, when a
regressor is used to map visual features to semantic embedding, the
disjoint training and testing classes in ZSL intrinsically require the
regressor to generalise out-of-bounds. This inherently limits the
accuracy of ZSL recognition. In this work, we address the issue of 
the generalisation capability of a ZSL mapping regressor from both the
model- and data-centric perspectives: (1) by proposing a more robust regression model
with better generalisation properties, and (2) improving model learning by 
augmenting auxiliary data  with a re-weighted additional dataset according to the relevance to the target problem.

\noindent\textbf{Multi-Task Embedding}\quad When
establishing the mapping between visual features and semantic
embeddings, most ZSL methods learn each dimension of this mapping
{\em independently} -- whether semantic embedding is discrete as in the
case of attributes \cite{Lampert2009,Akata2015}, or continuous as in
the case of word vectors \cite{Socher2013,FuXKG_CVPR15}. This strategy is  likely to overfit to the training
classes 
 because it treats each dimension of the label
in semantic embedding independently despite the labels living on a
non-uniform manifold \cite{mahadevan2015matrixManifold} and many independent
mappings result in a large number of parameters to be learned. We denote this
conventional approach as Single-Task Learning (STL) due to
the independent learning of mappings for each attribute/word
dimension. In contrast, we advocate a Multi-Task Learning (MTL)
\cite{Evgeniou2004,Kumar2012,Pan2010} regression approach to mapping
visual features and their semantic embeddings. By 
constraining the mapping parameters of each learning task to lie closely on a
low-dimensional manifold, we gain two advantages: (1) Exploiting the
relation between the response variables (dimensions of the label
embedding), (2) reducing the total number of parameters to
fit. The resulting visual-semantic mapping is more robust to the
domain shift between ZSL training and testing classes.  As a helpful
byproduct, the MTL mapping, provides a lower dimensional latent space
in which the nearest neighbour (NN) matching required by ZSL can be better performed
\cite{Beyer1999} compared to the usual higher dimensional label
semantic embedding space.

\begin{figure}[t!]\label{fig:NegativeAuxData}
\centering
\includegraphics[width=0.9\linewidth]{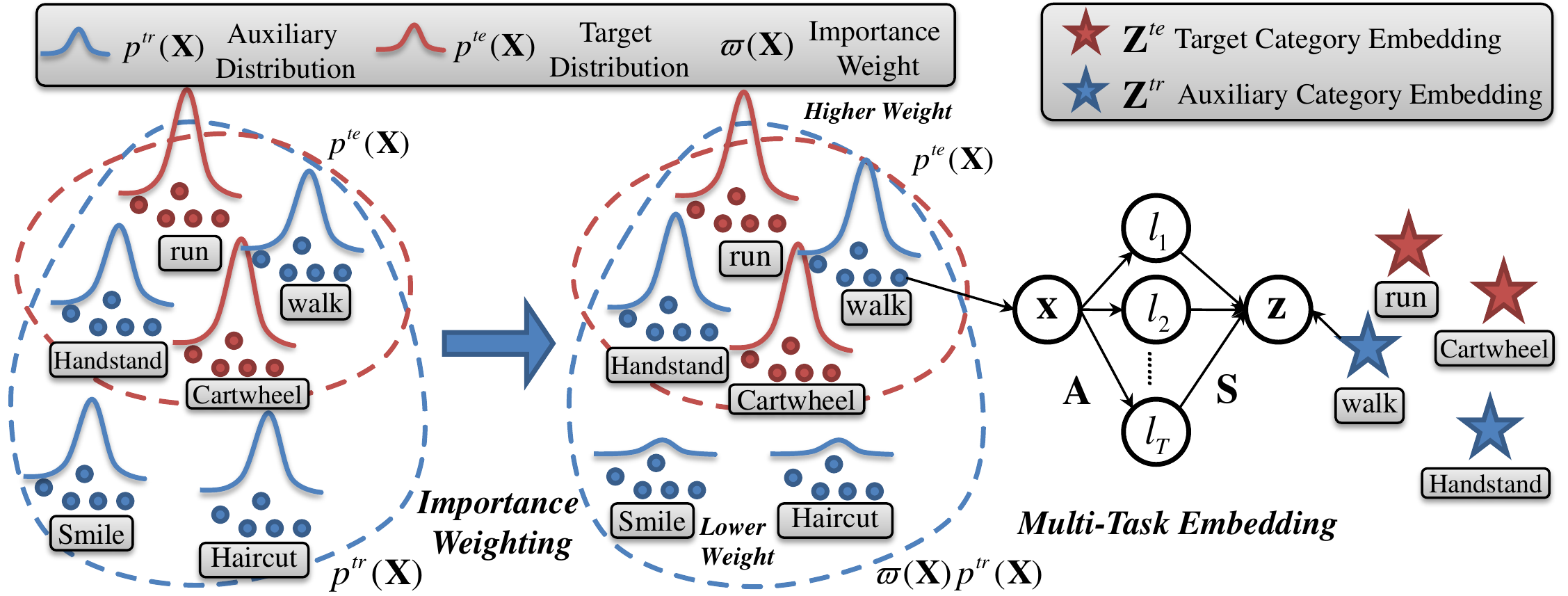}
\caption{\textcolor{black}{Two strategies to improve generalisation of
    visual-semantic mapping in ZSL. Left: Importance weighting to
    prioritise auxiliary data relevant to the target
    domain. Right: Learning the mapping from visual features $\matr{X}$ to
    semantic embedding $\matr{Z}$ by MTL reduces overfitting, and also
    provides a latent lower dimensional representation $\{\vect{l}_t\}$ to benefit nearest neighbour matching.}}
\end{figure}

\noindent\textbf{Prioritised Auxiliary Data Augmentation
  for Domain Adaptation}\quad \newline From a data-, rather than model-centric
perspective, studies have also attempted to improve the generalisation
of ZSL methods by augmenting\footnote{{In this work,
    data augmentation means exploiting additional data in a wider context
    from multiple data sources, in contrast to synthesising more
    artificial variations of one dataset as in deep learning.}}
the  auxiliary dataset with  additional datasets containing a  
wider array of classes and instances \cite{Xu2015,Habibian2014}. 
The idea is that including a broader additional set should provide better coverage of the visual feature and label embedding spaces, 
 therefore helping to learn a visual-semantic mapping that better generalises to target classes, and thus improves performance when representing and recognising target classes.
However, existing studies on exploring this idea have been
rather crude, e.g. simply expanding the training dataset by blindly
concatenating auxiliary set with additional data \cite{Xu2015}. This is not only inefficient but also
dangerous, because it does not take into account the (dis)similarity
between the extra incorporated data and the target classes for
recognition, thus risking {\em negative transfer} \cite{Pan2010}. In
this work, we address the issue that auxiliary and target
data/categories will have different marginal distributions (Fig~\ref{fig:NegativeAuxData}).
We selectively re-weight those relevant instances/classes from the auxiliary data  
that are expected to improve the the visual-semantic mapping in the context of the specific 
 target classes to be recognised (target
domain). We formulate this prioritised data augmentation as
a domain adaptation problem by minimizing 
the discrepancy between the marginal distributions of the auxiliary and
target domains. To achieve this, we propose an importance weighting
strategy to re-weight each auxiliary instance in order to minimise the
discrepancy. Specifically we generalise the classic
\textit{Kullback-Leibler Importance Estimation Procedure}
(KLIEP) \cite{Sugiyama2007,Garcke2014} to the zero-shot learning
problem. 

\section{Related Work}
\noindent\textbf{Zero-Shot Learning}\quad
Zero-shot Learning (ZSL) \cite{Lampert2009} aims to generalize
existing knowledge to recognize new categories without training
examples by re-using a mapping learned from visual features to their
semantic embeddings. Commonly used label embeddings are semantic
attributes \cite{Lampert2009,Liu2011,Fu2015} and word-vectors
\cite{Socher2013,Xu2015}. The latter has the advantage of being
learned from data without requiring manual annotation. Commonly used
visual-semantic mappings include linear \cite{dinu2014improving} and non-linear
regression \cite{Fu2015,Socher2013,Xu2015}, classification
\cite{Lampert2009,Liu2011}, and bilinear ranking \cite{Akata2015}.  

Existing ZSL methods suffer from weak generalisation due to the
domain-shift induced by disjoint auxiliary-target classes, an issue that
has recently been highlighted explicitly in the literature
\cite{FuXKG_CVPR15,Fu2015,dinu2014improving,Lazaridou2014}. Attempts to address this so far include post-processing heuristics
\cite{Fu2015,dinu2014improving,Lazaridou2014}, sparse coding regularisation
\cite{FuXKG_CVPR15}, and simple blind enlarging of the training set
 with auxiliary data \cite{Xu2015}.  In contrast to \cite{FuXKG_CVPR15,Xu2015}, we focus
on: (1) Building a visual-semantic mapping with intrinsically better
generalisation properties, and (2) re-weighting the auxiliary set to
prioritise  auxiliary instances most relevant to the target
instances and classes. Our method is complementary to \cite{Fu2015,dinu2014improving}
and can benefit from these heuristics.

\noindent\textbf{Zero-Shot Action Recognition}\quad
Among many ZSL tasks in computer vision, zero-shot action recognition
\cite{Liu2011,Xu2015,Kodirov2015,gan2016recognizing,chang2016dynamic} is of particular interest because of
the lesser availability of {\em labelled} video compared to image
data and videos are more difficult to label than static
images due to extended temporal duration and more complex ontology.
ZSL action recognition is much less studied than still
image recognition, and existing video-ZSL methods suffer from the same
domain-shift drawbacks highlighted above. 

\noindent\textbf{Multi-Task Regression Learning}\quad
Multi-Task Learning (MTL) \cite{Pan2010,Yang2015} aims to improve
generalisation in a set of supervised learning tasks by
modelling and exploiting  shared knowledge across the tasks.
An early study \cite{Evgeniou2004} proposed to model
the weight vector for each task $\scal{t}$ as a sum of a shared global task
$\vect{w}_0$ and task specific parameter vector
$\vect{w}_t$. However, the assumption of a globally shared
underlying task is too strong, and risks inducing {\em negative transfer}
\cite{Pan2010}. This motivates the Grouping and Overlapping
Multi-Task Learning (GOMTL)  \cite{Kumar2012} framework which instead assumes that each
task's weight vector is a task-specific combination of a small set
of latent basis tasks. This constrains the parameters of all tasks to
lie on a low dimensional manifold.

MTL methods have been studied for action recognition
\cite{Zhou2013,yuan2013multi,liu2015single,mahasseni2013latent}. However,
all of these studies focus on improving standard {\em supervised} action
recognition with multi-task sharing. For example, considering each of
multiple views \cite{liu2015single,mahasseni2013latent}, feature
modalities \cite{yuan2013multi}, or -- most obviously -- action
categories \cite{Zhou2013} as different tasks. Multi-view/multi-feature 
recognition is orthogonal to our work, while the later ones are
concerned with supervised recognition, and cannot be 
generalised to the ZSL scenario. In contrast, we take a very
different approach and treat each dimension of the visual-semantic
mapping as a task, in order to leverage MTL to improve auxiliary-target
generalisation across the disjoint target categories. Finally, we
note that the use of MTL to learn the visual semantic mapping provides
a further benefit of a lower-dimensional space in which zero-shot
recognition can be better performed due to being more meaningful for
NN matching \cite{Beyer1999}.

\noindent\textbf{Importance Weighting for Domain Adaptation}\quad
Domain shift is a widely studied problem in transfer learning
\cite{Pan2010}, although it is usually induced by sampling bias
\cite{torralba2011dataset_bias,Huang2007} or sensor change
\cite{saenko2010domainAdapt} rather than the disjoint categories in
ZSL.  Importance weighting (IW) \cite{Sugiyama2007,Huang2007} has been one
of the main adaptation techniques to address this issue. 
The prior work in this
area is designed for the standard domain transfer problem in a
{\em supervised} learning setting \cite{Pardoe2010}, while we are the first to generalise it to the
{\em zero-shot} learning scenario. The IW technique we generalise is related to 
another domain adaptation approach based on  discovering a feature mapping
to minimise the \textit{Maximum Mean Discrepancy} (MMD) \cite{gretton2006kernel,baktashmotlagh2013unsupDA} between distributions. 
However MMD, is less appropriate for us due to focus on feature mapping rather
than instance reweighing, and our expectation is that 
only subsets of auxiliary instances will be relevant to
the target rather than the holistic auxiliary set. 

\noindent\textbf{Contributions}\quad This paper
contributes both model- and data-centric strategies to improve
ZSL action recognition: (1) We formulate learning a more
generalisable visual-semantic mapping in ZSL as a multi-task learning
problem with a lower-dimensional latent semantic embedding space for
more effective matching. (2) We improve visual-semantic regression
generalisation by prioritised data augmentation using importance
weighting of auxiliary instances relevant to the target domain.

\begin{table}[t!]
\centering
\caption{Notation Summary}
\label{tab:Notations}
\resizebox{0.95\linewidth}{!}{ 
\begin{tabular}{ll}
\hline
\textbf{Notation}                   & \textbf{Description}                     \\ \hline
$n^{tr}_c; n^{te}_c$					&	Number of training categories ; testing categories	\\
$n^{tr}_x; n^{te}_x$						& Number of all training instances; all testing instances	\\
$\matr{X}\in \mathbb{R}^{d_x\times n_x}$; $\vect{x}_i$     & Visual feature matrix for N instances; column representing the $i$-th instance   \\
$\matr{Y}\in \{0,1\}^{n_c\times n_x}$; $\vect{y}_i$      & Binary class labels for N instances 1-of-$n_c$ encoding; column representing the $i$-th instance      \\
$\matr{V}\in \mathbb{R}^{d_z\times n_c}$;     & Semantic label embedding for $n_c$ categories;     \\
$\matr{Z}\in \mathbb{R}^{d_z\times n_x}$; $\vect{z}_i$     & Semantic label embedding for $n_x$ instances; column representing the $i$-th instance       \\
$\matr{W}\in \mathbb{R}^{d_z\times d_x}$; $\vect{w}_d$	&	STL regression coefficient matrix; row representing the regressor for the $d$-th dimension	\\
$\matr{A}\in \mathbb{R}^{T\times d_x}$ ; $\vect{a}_t$    & MTL regression coefficient matrix; row representing the regressor for the $t$-th latent task            \\
$\matr{S} \in \mathbb{R}^{d_z\times T}$; $\vect{s}_d$		& MTL linear combination matrix; row representing linear combination vector for the $d$-th output \\
$\matr{L}\in \mathbb{R}^{T\times n_x}$; $\vect{l}_i$	& Latent space embedding for visual instances; column is $i$th instance\\
$\vect{\omega} \in \mathbb{R}^{{n_x}\times 1}$	& weighting vector for auxiliary data \\
$f: \matr{X} \to \matr{Z}$                          & Visual to semantic mapping function      \\
 \hline
\end{tabular}}
\end{table}

\section{Visual-Semantic Mapping with Multi-Task Regression}

In ZSL, we aim to recognise action
categories $\matr{Y}$ given visual features $\matr{X}$ where training/auxiliary and
testing/target categories do not overlap $\set{Y}^{tr}\cap
\set{Y}^{te}=\emptyset$. The key method by which ZSL is achieved
is to embed each category label in $\set{Y}$ into a semantic label embedding space $\set{Z}$ which provide a vector representation of any
\emph{nameable} category. Table~\ref{tab:Notations} summarises the notation used
in the subsequent sections.

\subsection{Training a Visual Semantic Mapping}\label{sec:trainMapping}
We first introduce briefly the conventional
single task learning using regression for visual-semantic mapping\cite{dinu2014improving,Xu2015,Fu2015}.

\noindent\textbf{Single-Task Regression}\quad\label{sect:WVEmbedding}
Given a matrix $\matr{V}$ describing the embedded action
names\footnote{To deal with multi-word compound action category names,
  e.g. ``Apply Eye Makeup", we apply a simple average, 
  summing the component word vectors \cite{Xu2015,Fu2015}.}, and
per-video binary labels $\matr{Y}$, we firstly obtain the label embedding of any
action label for a video clip as $\vect{z}_i=\matr{V}\vect{y}_i$. We then learn a
visual-semantic mapping function $f:\set{X}\to\set{Z}$ on the
training categories.  Given a loss function $l(\cdot,\cdot)$, we learn
the mapping $f$ by optimising Eq~(\ref{eq:KernelRidgeRegression})
where $\Omega(f)$ denotes  regularization on the mapping:

\begin{equation}\label{eq:KernelRidgeRegression}
\resizebox{.4\hsize}{!}
{$
\begin{split}
\min\limits_{f}\frac{1}{n^{tr}_x}\sum\limits_{i=1}^{n^{tr}_x} l\left(f(\mathbf{x}_i),\mathbf{z}_{i}\right)+\Omega(f).\\
\end{split}
$}
\end{equation}

\noindent The most straightforward choice of mapping $f$ and loss $l$ is
linear $f(\vect{x})=\matr{W}\vect{x}$, and square error respectively,
which results in a regularized linear (ridge) regression problem:
$l\left(f(\mathbf{x}_i),\mathbf{z}_{i}\right)=||\mathbf{z}_i-\matr{W}\mathbf{x}_i||_2^2$.
A closed-form solution to $\matr{W}$ can then be obtained by $\matr{W} = \matr{Z}\matr{X}^T\left(\matr{XX}^T+\lambda n_x^{tr}\matr{I}\right)^{-1}$. Each row $\mathbf{w}_d$ of regressor $\matr{W}$ maps visual feature $\mathbf{x}_i$ to $d$th dimension of response variable $\mathbf{z}_i$. Since regressors $\{\mathbf{w}_d\}_{d=1\cdots d_z}$ are learned independently from each other this is referred as \textbf{single-task learning (STL)} with each $\mathbf{w}_d$ defining one distinct `task'.

\noindent\textbf{From Single to Multi-Task Regression}\quad\label{sect:ReducedRankEmbedding}
 In the conventional ridge-regression solution to
 Eq.~(\ref{eq:KernelRidgeRegression}), each task
$\mathbf{w}_d$ is effectively learned
 separately, ignoring any relationship between tasks. We wish to model
 this relationship by discovering a latent basis of 
 predictors such that tasks $\mathbf{w}_d$ are constructed as
 linear combinations of $T$ latent tasks $\{\mathbf{a}_t\}_{t=1\cdots
   T}$. So the $d$th regression predictor is now modelled as
 $\mathbf{w}_d=\sum_t{s}_{dt}\mathbf{a}_t=\mathbf{s}_d^T\matr{A}$, where
 $\mathbf{s}_d$ is the combination coefficient for $d$-th
 task. Denoting  multi-task regression prediction as
 $f(\mathbf{x}_i,\matr{S},\matr{A})$, we now optimise:
 
\begin{equation}\label{eq:MTL_Loss}
\resizebox{.6\hsize}{!}
{$
\min\limits_{\matr{S},\matr{A}} \frac{1}{n^{tr}_x}\sum\limits_{i=1}^{n^{tr}_x}l(f(\mathbf{x}_i,\matr{S},\matr{A}),\mathbf{z}_i) + \lambda \Omega(S) + \gamma \Psi(\matr{A}).
 $}
\end{equation}

\noindent\textbf{Grouping and Overlap Multi-Task
  Learning}\quad An effective method following the MTL design pattern
above is GOMTL \cite{Kumar2012}. GOMTL uses a $\matr{W}=\matr{SA}$ task parameter
matrix factorisation, where the number of latent tasks $T$ (typically
$T< d_z$) is a free parameter. Requiring the combination coefficients
$\mathbf{s}_t$ to be sparse, via a $\ell_1$ regulariser, the loss is
written as  

\begin{equation}
\resizebox{.7\hsize}{!}
{$
\min\limits_{\{\mathbf{s}_t\},\matr{A}} \sum\limits_{t=1}^T\frac{1}{n^{tr}_x}\sum\limits_{i=1}^{n^{tr}_x}||\vect{z}_{t,i}-\mathbf{s}_t\matr{A}\mathbf{x}_i||+\lambda\sum_{t=1}^T||\mathbf{s}_t||_1 + \gamma||\matr{A}||_F^2 \label{eq:GOMTL}
$}
\end{equation}

This  can be solved by iteratively updating $\matr{A}$ and $\matr{S}$. When $\matr{A}$ is fixed, loss function reduces to
 a standard L1 regularized (LASSO) regression  problem that can be efficiently solved by Alternating Direction Method of Multipliers (ADMM) \cite{boyd2011distributed}.
When $\matr{S}$ is fixed, we can efficiently solve $\matr{A}$ by gradient descent. 



\noindent\textbf{Regularized Multi-Task Learning (RMTL)} \quad
The classic RMTL method \cite{Evgeniou2004} models task parameters as
the sum of a globally shared and task specific parameter vector:
$\textbf{w}_t=\textbf{a}_0+\textbf{a}_t$. It can be seen that this
corresponds to a special case of GOMTL's $\matr{W}=\matr{SA}$ predictor matrix
factorisation \cite{Yang2015}. Here there are  $T=d_z+1$ latent tasks, a fixed
task combination vector $\mathbf{s}_t = [1 \quad
  \mathbf{1}(t=1) \quad \mathbf{1}(t=2) \cdots \mathbf{1}(t=d_z)]^T$
where $\mathbf{1}(\cdot)$ is the indicator function and ${A}=\left[\mathbf{a}_0^T \mathbf{a}_1^T \cdots
\mathbf{a}_{d_{z}}^T\right]^T$.

\noindent\textbf{Explicit Multi-Task Embedding (MTE)}\quad
In GOMTL Eq~(\ref{eq:GOMTL}), it can be seen that the label embedding
$\mathbf{z}_i$ is approximated from the data by the mapping
$\mathbf{s}_t\matr{A}\mathbf{x}_i$, and this approximation is reached by
combination via the latent representation $\matr{A}\mathbf{x}_i$. While GOMTL defines
this space implicitly via the learned $\matr{A}$, we propose to model it
explicitly as $\mathbf{l}_i\approx \matr{A}\mathbf{x}_i$. This is so the
actual projections $\mathbf{l}_i$ in this latent space can be
regularised explicitly, in order to learn a latent space which
generalises better to test data, and hence improves ZSL matching
later. 

Specifically, we split the GOMTL loss $||\mathbf{z}_i-\matr{SA}\mathbf{x}_i||^2_2$ into two parts:  $||\mathbf{l}_i-\matr{A}\mathbf{x}_i||^2_2$ and $||\mathbf{z}_i-\matr{S}\mathbf{l}_i||^2_2$ to learn the mapping to the latent space, and from the latent space to the label embedding respectively. This allows us to place additional regularization on $\mathbf{l}_i$ to avoid extreme values in the latent space and thus later improve neighbour matching (Section~\ref{sec:NNMatching}). Given the large and high dimensional video datasets, we  apply Frobenius norm on  $\matr{S}$ in contrast to GOMTL's $\ell_1$.

\begin{equation}\label{eq:VideoStoryLoss}
\resizebox{.65\hsize}{!}
{$
\begin{split}
\min\limits_{\{\mathbf{s}_t\},\matr{A},\{\mathbf{l}_i\}}&\quad\sum\limits_{t=1}^T\frac{1}{n^{tr}_x}\sum\limits_{i=1}^{n^{tr}_x}\left(||\vect{z}_{t,i}-\mathbf{s}_t\mathbf{l}_i||^2_2+||\mathbf{l}_i-\matr{A}\mathbf{x}_i||^2_2\right)+\\
&\lambda_S\sum\limits_{t=1}^T||\mathbf{s}_t||_2^2 + \lambda_A||\matr{A}||_F^2 +\lambda_L\sum\limits_{i=1}^{n_x^{tr}}||\mathbf{l}_i||_2^2\\
\end{split}
$}
\end{equation}

\noindent Our explicit multi-task embedding has similarities to
\cite{Habibian2014}, but our purpose is multi-task regression for ZSL,
rather than embedding for video descriptions. To solve our explicit
embedding model we iteratively solve $\matr{L}$,$\matr{A}$ and $\matr{S}$ while fixing the
other two. With the $\ell_2$ norm on $\matr{S}$, this has a convenient
closed-form solution to each parameter: 

\textcolor{black}{\begin{equation}
\resizebox{.45\hsize}{!}
{$
\begin{split}
&\matr{L}=(\matr{S}^T\matr{S}+(\lambda_L n^{tr}_x + 1)\mathbf{I})^{-1}(\matr{S}^T\matr{Z}+\matr{AX})\\
&\matr{S}=\matr{ZL}^T(\matr{LL}^T+\lambda_S n^{tr}_x \mathbf{I})^{-1}\\
&\matr{A}=\matr{LX}^T(\matr{XX}^T+\lambda_A n^{tr}_x \mathbf{I})^{-1}\\
\end{split}
$}
\end{equation}}

\subsection{Zero-Shot Action Recognition}\label{sec:NNMatching}
We consider two alternative NN matching
methods for zero-shot action prediction that use the MTL mappings described above. 

\noindent\textbf{Distributed Space Matching}\quad
Given a trained visual-semantic regression $f$, we project testing set visual feature $\vect{x}^{te}$ into the semantic label embedding space. The standard strategy \cite{Xu2015,Fu2015,dinu2014improving} is then to employ NN matching in this space for zero-shot recognition. Specifically, given the matrix of label embeddings for each target category name $\matr{V}^{te}$, and using cosine distance norm, the testing video $\vect{x}^{te}$ are classified by:


\begin{equation}\label{eq:stlMatching}
\resizebox{.4\hsize}{!}
{$
\matr{y}^*=arg\min\limits_{\matr{y}^*} ||\matr{V}^{te}\matr{y}^*-f(\matr{x}^{te})||
$}
\end{equation}

\noindent where $f(\matr{x}^{te})=\matr{Wx}^{te}$ for STL and $f(\matr{x}^{te})=\matr{SAx}^{te}$ for MTL.

\noindent\textbf{Latent Space Matching}\quad
MTL methods provide an alternative to matching in label space:
Matching in the latent space. The representation of testing data
in this space is the output of latent regressors
$\vect{l}_{te}=\matr{A}\vect{x}^{te}$ (Eq.~(\ref{eq:VideoStoryLoss})). To get the
representation of testing categories in the latent space we invert the
combination matrix $\matr{S}$ to project target category names $\matr{V}^{te}$ into
latent space. Specifically we classify by Eq.~(\ref{eq:LatentMatching}), where
$(\matr{S}^T\matr{S})^{-1}\matr{S}^T$ is the Moore-Penrose pseudoinverse. 

\begin{equation}\label{eq:LatentMatching}
\resizebox{.52\hsize}{!}
{$
\vect{y}^*=arg\min\limits_{\vect{y}^*} ||(\matr{S}^T\matr{S})^{-1}\matr{S}^T\matr{V}^{te}\vect{y}^*-\matr{AX}^{te}||
$}
\end{equation}

\noindent NN matching in the latent space is better than in semantic label space because: (i) the dimension is lower $T<d_z$, and (ii) we have explicitly regularised the latent space to be well behaved (Eq.~(\ref{eq:VideoStoryLoss})).


\section{Importance Weighting}\label{sect:ImportanceWeighting}


Augmenting auxiliary data with additional examples from other datasets has been proved to benefit learning the visual-semantic mapping \cite{Xu2015}.
However, simply aggregating auxiliary and additional datasets is not ideal as including irrelevant data risks `negative transfer'. Therefore we are motivated to develop methodology to prioritise augmented auxiliary data that is useful for a particular ZSL recognition scenario.  Specifically, we  learn a per-instance weighting $\omega(\vect{x})$ on the auxiliary dataset $\matr{X}^{tr}$ to adjust each instance's contribution according to relevance to the target domain. 
Because Importance Weighting (IW) adapts auxiliary data to the target domain, we assume a transductive setting with access to testing data $\matr{X}^{te}$.

\noindent\textbf{Kullback-Leibler Importance Estimation Procedure (KLIEP)}\quad
We first introduce the way to estimate a per-instance auxiliary-data weight given the distribution of target data $\matr{X}^{te}$. This is based on the idea \cite{Sugiyama2007} of minimizing the KL-divergence ($D_{KL}$) between training $p^{tr}(\matr{x})$ and testing data distribution $p^{te}(\matr{x})$ via learning a weighting function $\omega(\matr{x})$. This is formalised in Eq.~(\ref{eq:KLIEPX_Obj}):

\begin{equation}\label{eq:KLIEPX_Obj}
\resizebox{.65\hsize}{!}
{$
\begin{split}
&\min\limits_{\omega} D_{KL}(p^{te}(\matr{x})| \omega(\matr{x})p^{tr}(\matr{x}))=\int p^{te}(\matr{x})\log \frac{p^{te}(\matr{x})}{\omega(\matr{x})p^{tr}(\matr{x})}d\matr{x}\\
&\min\limits_{\omega}\int p^{te}(\matr{x})\log \frac{p^{te}(\matr{x})}{p^{tr}(\matr{x})} d\vect{x} - \int p^{te}(\matr{x})\log \omega(\matr{x}) d\matr{x}
\end{split}
$}
\end{equation}

\noindent The first term is fixed w.r.t. $\omega(\matr{x})$ so the objective to optimise is:


\begin{equation}
\begin{split}
&\min\limits_{\omega}- \int p^{te}(\matr{x})\log \omega(\matr{x}) d\matr{x} \approx -\frac{1}{n_x^{te}}\sum\limits_{i=1}^{n_x^{te}}\log \omega(\mathbf{x}_i)
\end{split}
\end{equation}


\noindent\textbf{Aligning Both Visual Features and Labels}\quad
KLIEP is conventionally used for domain adaptation by reweighting instances \cite{Sugiyama2007,Pardoe2010}. In the case of transductive ZSL, we have the target data $\matr{X}^{te}$ and category labels $\matr{Z}^{te}$ respectively, although not instance-label association which is to be predicted. In this case we can further improve ZSL by extending KLIEP to align training and testing sets in both visual feature and category sense\footnote{KLEIP with labels was studied by \cite{Garcke2014}, but they assumed the target joint distribution of $\matr{X}$ and $\matr{Z}$ is known. So \cite{Garcke2014} is only suitable for traditional supervised learning with labeled target examples of $\mathbf{z}_i$ and $\mathbf{x}_i$ in correspondence. In our case we have the videos to classify and the zero-shot category names, but the assignment of names to videos is our task rather than prior knowledge.}. Specifically, we minimise the kullback-leibler divergence between the target and auxiliary in terms of both the visual and category distributions: 

\begin{equation}
\resizebox{.75\hsize}{!}
{$
\begin{split}
&\min\limits_{\omega_x,\omega_z}D_{KL}(p^{te}(X)|| \omega_x(\matr{X})p^{tr}(\matr{X}))+D_{KL}(p^{te}(\matr{Z})||\omega_z(\matr{Z})p^{tr}(\matr{Z}))\\
&\min\limits_{\omega_x,\omega_z} -\frac{1}{n_x^{te}}\sum\log \omega_x(\mathbf{x}_i^{te}) -\frac{1}{n_x^{te}}\sum\log \omega_z(\mathbf{z}_i^{te})
\end{split}
$}
\end{equation}

Given both $\matr{X}^{te}$ and $\matr{Z}^{te}$, we construct the weighting functions as a combination of  Gaussian kernels centered at the testing data and categories. Specifically we define $\omega(\mathbf{x},\mathbf{z})=\omega_x(\mathbf{x})+\omega_z(\mathbf{z})$ where $\omega_x(\mathbf{x})$ and $\omega_z(\mathbf{z})$ are calculated as in Eq.~(\ref{eq:GaussianKernel}).  Here $\omega(\mathbf{x},\mathbf{z})$ extends the previous notation $\omega(\mathbf{x})$ to indicate giving a weight to each training instance given visual feature $\mathbf{x}$ and class name embedding $\mathbf{z}$. So if there are $n^{tr}_x$ instances, $\omega(\mathbf{x},\mathbf{z})$ returns a weight vector of length $n^{tr}_x$.


\begin{equation}\label{eq:GaussianKernel}
\resizebox{.9\hsize}{!}
{$
\begin{split}
\omega_x(\mathbf{x})=\sum\limits_{i=1}^{n^{te}_x}\alpha_i \phi(\mathbf{x},\mathbf{x}_i^{te}), \quad
\omega_z(\mathbf{z})=\sum\limits_{i=1}^{n^{te}_x}\beta_j \phi(\mathbf{z},\mathbf{z}_i^{te}), \quad \phi(\mathbf{x},\mathbf{x}^{te}_i)=exp\left(-\frac{||\mathbf{x}-\mathbf{x}^{te}_i||^2}{2\sigma^2}\right)
\end{split}
$}
\end{equation}

\noindent For  ease of formulation, we denote $\mathbf{a}=[\alpha_1 \cdots \alpha_{n^{te}_x}]^T$, $\mathbf{b}=[\beta_1 \cdots \beta_{n^{te}_x}]^T$, $\Phi_{\mathbf{a}}(\mathbf{x})=[\phi(\mathbf{x},\mathbf{x}_1^{te}) \cdots \phi(\mathbf{x},\mathbf{x}_{n^{te}_x}^{te})]^T$ and $\Phi_{\mathbf{b}}(\mathbf{z})=[\phi(\mathbf{z},\mathbf{z}_1^{te}) \cdots \phi(\mathbf{z},\mathbf{z}_{n^{te}_x}^{te})]^T$. The optimization can be thus written as

\begin{equation}\label{eq:XY_Prior_KLIEP}
\resizebox{.91\hsize}{!}
{$
\begin{split}
&\min\limits_{\mathbf{a},\mathbf{b}} -\frac{1}{n_x^{te}}\sum\limits_{i=1}^{n_x^{te}}\log \mathbf{a}^T\Phi_\mathbf{a}(\mathbf{x}_i^{te}) -\frac{1}{n_x^{te}}\sum\limits_{i=1}^{n_x^{te}}\log \mathbf{b}^T\Phi_\mathbf{b}(\mathbf{z}_i^{te}), \quad s.t. \quad \frac{1}{n_x^{tr}}\sum\limits_{i=1}^{n_x^{tr}}\omega(\mathbf{x}_i^{tr},\mathbf{z}_i^{tr}) = 1
\end{split}
$}
\end{equation}

\noindent The above constrained optimization problem is convex
w.r.t. both $\mathbf{a}$ and $\mathbf{b}$. It
 can be solved by interior point methods using the
derivatives in Eq.~(\ref{eq:DerivativeL}):

\begin{equation}\label{eq:DerivativeL}
\resizebox{.8\hsize}{!}
{$
\begin{split}
&\nabla\mathbf{a}=-\frac{1}{n_x^{te}}\sum\limits_{i=1}^{n_x^{te}}\frac{1}{\mathbf{a}^T\Phi_\mathbf{a}(\mathbf{x}_i^{te})}\Phi_\mathbf{a}(\mathbf{x}_i^{te}), \quad \nabla\mathbf{b}=-\frac{1}{n_x^{te}}\sum\limits_{i=1}^{n_x^{te}}\frac{1}{\mathbf{b}^T\Phi_\mathbf{b}(\mathbf{z}_i^{te})}\Phi_\mathbf{b}(\mathbf{z}_i^{te})
\end{split}
$}
\end{equation}

\noindent\textbf{Weighted Visual-Semantic Regression}\quad
Given per-instance weights $\omega$ estimated above, we can rewrite the loss function for both single-task ridge regression and multi-task regression in Sec~\ref{sec:trainMapping} as \textcolor{black}{$\omega_il(f(\mathbf{x}_i,\matr{A}),\mathbf{z}_i)$ and $\omega_il(f(\mathbf{x}_i,\matr{S},\matr{A}),\mathbf{z}_i)$} respectively. All our loss functions have quadratic form, so the weight can be expressed inside the quadratic loss e.g. $\omega_i||\mathbf{z}_i-\matr{W}\mathbf{x}_i||^2_2=||\mathbf{z}_i\sqrt{\omega_i}-\matr{W}\mathbf{x}_i\sqrt{\omega_i}||_2^2$. Thus to incorporate the weight information we simply replace the original semantic embedding matrix with $\tilde{\mathbf{z}}_i=\mathbf{z}_i\sqrt{\omega_i}$ and data matrix with $\tilde{\mathbf{x}}_i=\mathbf{x}_i\sqrt{\omega_i}$.


\section{Experiments}

\noindent\textbf{Datasets and Settings}\quad We evaluated our
contributions on three human action recognition datasets, HMDB51
\cite{Kuehne2011}, UCF101 \cite{Soomro2012} and Olympic Sports
\cite{Niebles2010}.  They contain 6766, 13320, 783 videos and 51, 101,
16 categories respectively. For all datasets we extract improved
trajectory feature (ITF) \cite{Wang2014}, a state-of-the-art
space-time feature representation for action recognition. We use Fisher Vectors (FV)
\cite{Perronnin2010} to encode three raw descriptors (HOG, HOF and
MBH). Each descriptor is reduced to half of its original dimension by
PCA, resulting in a 198 dim representation. Then we randomly sample
256,000 descriptors from all videos and learn a Gaussian Mixture with
128 components to obtain the FVs. The final dimension of FV encoded
feature is $2\times128\times198=50688$ dimensions. For the
label-embedding, we use 300-dimensional word2vec
\cite{Mikolov2013a}. We use $T=n^{tr}_c$ latent tasks, and
cross-validation to determine regularisation strength hyper-parameters
for the models\footnote{Ridge Regression (RR) has $15$M
  ($300\!\times\!50688$) parameters, whilst for HMDB51 where
  $T\!=\!25$, GOMTL and MTE have $1.27$M ($50688\!\times\!25 \!+\! 25\!\times\!300$)
  parameters. }.


\subsection{Visual-semantic Mappings for Zero-Shot Action Recognition}

\noindent\textbf{Evaluation Criteria}\quad To evaluate
zero-shot action recognition, we  divide each dataset evenly into
training and testing parts with 5 random
splits. Using classification accuracy for HMDB51 and UCF101 and
average precision for Olympic Sports as the evaluation metric, the
average and  standard deviation over the 5 splits are reported for
each dataset. 

\noindent\textbf{Compared Methods}\quad We study the
efficacy of our contributions by evaluating the different
visual-semantic mappings presented in Sec~\ref{sec:trainMapping}. We compare MTL-regression
methods with conventional STL Ridge Regression (denoted
\textbf{RR}) for ZSL. For RR/STL, nearest neighbour
matching is used to recognise target categories. Note that the RR+NN
method here corresponds to the core strategy used by
\cite{Xu2015,Fu2015,dinu2014improving}. 
The multi-task models we explore include: \textbf{RMTL} \cite{Evgeniou2004}: assumes each task's predictor is the sum of a global latent vector and a task-specific vector.
\textbf{GOMTL} \cite{Kumar2012}: Uses a predictor-matrix factorisation assumption in which tasks' predictors  lie on a low-dimensional subspace. \textbf{Multi-Task Embedding (MTE)}: \quad Our model differs from GOMTL in that it explicitly models and regularises a lower dimensional latent space. 
For the multi-task methods, we also compare the ZSL matching
strategies introduced in Section~\ref{sec:NNMatching}:
\textbf{Distributed:} Standard NN matching (Eq.~(\ref{eq:stlMatching})), and  \textbf{Latent:} our
proposed latent-space matching (Eq.~(\ref{eq:LatentMatching})).

\begin{table}[t]
\centering
\caption{Visual-semantic mappings for zero-shot action recognition: MTL (\checkmark) versus  STL (\text{\sffamily X}). Latent matching (\checkmark) versus distributed (\text{\sffamily X}) matching}
\label{tab:MTL_Perf}
\resizebox{0.65\linewidth}{!}{
\begin{tabular}{l|c|c|c|c|c}
\toprule
\specialcell{ZSL Model}   & MTL    & \specialcell{Latent Matching}  & HMDB51 &  UCF101  & \specialcell{Olympic Sports}  \\ \hline

RR           & \text{\sffamily X} & NA & $18.3\pm 2.1$ & $ 14.5\pm 0.9$ & $ 40.9 \pm 10.1$       \\
 \hline
RMTL \cite{Evgeniou2004}                  & \checkmark           & \text{\sffamily X} & $18.5\pm 2.1$ & $ 14.6\pm 1.1$ & $41.1 \pm 10.0$        \\
RMTL \cite{Evgeniou2004}   & \checkmark           & \checkmark           & ${18.7\pm 1.7}$ & $ 14.7\pm 1.0$ & $41.1 \pm 10.0$        \\
GOMTL \cite{Kumar2012}                       & \checkmark           & \text{\sffamily X} & $18.5\pm 2.2$ & $ 13.1\pm 1.5$ & $43.5\pm 8.8$         \\
GOMTL \cite{Kumar2012} & \checkmark     & \checkmark           & $18.9\pm 1.0$ & $ {14.9\pm 1.5}$ & $44.5\pm 8.5$         \\ 
MTE & \checkmark           & \text{\sffamily X} &   $18.7\pm 2.2$  &   $14.2\pm 1.3$  &                $\mathbf{44.5\pm 8.2}$       \\
MTE & \checkmark           & \checkmark           &  $\mathbf{19.7\pm 1.6}$  &   $\mathbf{15.8\pm 1.3}$   &  ${44.3\pm 8.1}$        \\
\bottomrule
\end{tabular}}
\end{table}

\noindent\textbf{Results:}\quad The comparison of single
task ridge regression with our multi-task methods is presented in
Table~\ref{tab:MTL_Perf}. 
From these results we make the following
observations: (i) Overall our multi-task methods improve on the corresponding
single-task baseline of RR. MTL regression (RMTL, GOMTL and MTE) improves single-task ridge regression by $5-10\%$ in
relative terms, with the biggest margins visible on the Olympic Sports
dataset. (ii) Within multi-task models, the GOMTL with sparse $\ell_1$
regularization outperforms RMTL. This suggests learning the task
combination $\matr{S}$ from data is better than fixing it as in RMTL. (iii)
Our MTE generally outperforms other multi-task methods supporting the
explicit modelling and regularisation of the latent space. (iv) In
most cases, NN matching in the latent space improve zero-shot
performance. This is likely due to the lower dimension of the latent
space compared to the dimension of the original word vector
embedding, making NN matching more meaningful \cite{Beyer1999}. 


\subsection{Importance Weighted Data Augmentation}

We next evaluate the impact of importance weighting in data augmentation for zero-shot action recognition. We perform the same 5 random split benchmark for each dataset. For data augmentation, we augment each dataset's training split with the data from all other datasets. For instance, for ZSL on HMDB51 we augment the training data with all videos from UCF101 and Olympic Sports. 

\noindent\textbf{Compared Methods}\quad
We study the impact of the data augmentation methods:
\textbf{Naive DA:} Naive Data Augmentation  \cite{Xu2015,xu2015zero} simply assigns equal weight to each auxiliary training sample. 
\textbf{Visual KLIEP:} The auxiliary data is aligned with the testing sample distribution $\matr{X}^{te}$ (Eq.~(\ref{eq:KLIEPX_Obj})).
\textbf{Category KLIEP:} The auxiliary categories are aligned with testing category distribution $\matr{Z}^{te}$. This is achieved by the same prodcedure in Eq.~(\ref{eq:KLIEPX_Obj}) by replacing $\mathbf{x}$ with $\mathbf{z}$.
\textbf{Full  KLIEP:} The distribution of both samples $\matr{X}^{te}$ and categories $\matr{Z}^{te}$ is used to reweight the auxiliary data (Eq.~(\ref{eq:XY_Prior_KLIEP})).

\noindent\textbf{Results:}\quad  From the results in Table~\ref{tab:ImportanceWeighting}, we draw the
 conclusions: (i) Both the  baseline single task learning (STL) method and our
Multi-Task Embedding (MTE) improve with Naive DA (compare unaugmented results
in Table~\ref{tab:MTL_Perf}), (ii) The Visual, Category, and
Full visual+category-based weightings all improve on Naive DA in the case of STL RR. 
(iii) We see that our MTE with Full KLIEP augmentation performs the best overall. The ability of
KLIEP to improve on Naive DA suggests that the auxiliary data is
indeed of variable relevance to the target data, and selectively
re-weighing the auxiliary data is important. (iv) For KLIEP-based DA,
either Visual or Category DA provides most of the improvement, with
relatively less improvement obtained by using both together. 

\begin{table}[t]
\centering
\caption{Data augmentation and importance weighting for ZSL action recognition.}
\label{tab:ImportanceWeighting}
\resizebox{0.58\linewidth}{!}{
\begin{tabular}{cc|c|c|c}
\toprule
\multicolumn{1}{c|}{ZSL Model} & Weighting Model & HMDB51        & UCF101        & OlympicSports           \\ \hline
\multicolumn{1}{c|}{RR }               & Naive DA        & $21.9\pm 2.4$ & $19.4\pm 1.7$ & $46.5\pm 9.4$  \\
\multicolumn{1}{c|}{MTE}  & Naive DA        & $\mathbf{23.4\pm 3.4}$ & $\mathbf{20.9\pm 1.5}$ & $\mathbf{49.4\pm 8.8}$  \\\hline
\multicolumn{1}{c|}{RR}               & Visual KLIEP         & $23.2\pm 2.7$ & $20.3\pm 1.6$ &   $47.2\pm 9.3$                \\
\multicolumn{1}{c|}{RR}               & Category KLIEP         & $23.0\pm 2.1$ & $20.2\pm 1.6$ & $51.8\pm 8.7$                 \\
\multicolumn{1}{c|}{RR}               & Full KLIEP        & $23.7\pm 2.7$ & $20.7\pm 1.4$ &  $51.3\pm 9.0$                \\ 
\multicolumn{1}{c|}{MTE} & Visual KLIEP        & ${23.4\pm 2.8}$ & ${20.8\pm 2.0}$ &  ${51.4\pm 9.2}$               \\
\multicolumn{1}{c|}{MTE} & Category KLIEP        & ${23.3\pm 2.4}$ & ${20.9\pm 1.7}$ &  ${50.9\pm 8.3}$               \\

\multicolumn{1}{c|}{MTE} & Full KLIEP        & $\mathbf{23.9\pm 3.0}$ & $\mathbf{21.9\pm 2.7}$ &  $\mathbf{52.3\pm 8.1}$               \\
\bottomrule
\end{tabular}}
\end{table}

\noindent\textbf{Alternative Models}\quad We also compare against previous state-of-the-art methods including those driven by both attributes and word-vector category embeddings.
\textbf{DAP/IAP} \cite{Lampert2009}: Direct/Indirect attribute prediction are classic attribute-based zero-shot recognition models based on training SVM classifiers independently for each attribute, and using a probabilistic model to match attribute predictions with target classes.
\textbf{HAA}: We implement a simplified version of the Human Actions by Attributes model \cite{Liu2011}: We first train attribute detection SVMs, and test samples are assigned to categories based on cosine distance between their vector of attribute predictions and the target classes' attribute vectors. {\textbf{SVE} \cite{Xu2015}: Support vector regression was adopted to learn the visual to semantic mapping.} \textbf{ESZSL} \cite{Romera-paredes2015}: Embarrassingly Simple Zero-Shot Learning defines the loss function as the mean square error on label prediction in contrast to the regression loss defined in other baseline models. 
{\textbf{SJE}: Structured Joint Embedding \cite{Akata2015} employed a triplet hinge loss. The objective is to enforce relevant labels having higher projection values from visual features than those of non-relevant labels.} \textbf{UDA}: The Unsupervised Domain Adaptation model \cite{Kodirov2015} learns dictionary on auxiliary data and adapts it to the target data as a constraint on the target dictionary rather than blindly using the same dictionary. 
This work combines both attribute and word vector embeddings.

\noindent\textbf{Comparison Versus State of the
  Art:}\quad  Table~\ref{tab:ImportanceWeighting2} compares our models
with various contemporary and state-of-the-art models. For clear comparison, 
we indicate for each method which embedding
((\textbf{W})ordvector / (\textbf{A})ttribute) and feature (our FV, or BoW) are used,
as well as whether it has a transductive dependency on the test data (\textbf{TD}) or
exploits additional augmenting data (\textbf{Aug}). From these results we conclude that:
(i) Although data augmentation has a big impact, our
non-transductive and no data augmentation method ({MTE}) generally
outperforms prior alternatives due to learning an effective latent
matching space robust to the train/test class shift; {(ii) The performance of our MTE with word-vector embedding is strong when compared with DAP/IAP/HAA/ESZSL even with attribute embedding. Given the same attribute embedding, MTE outperforms all state-of-the-art models due to the discovery of latent attributes from the original attribute space}; (iii)  
Moreover, given importance weighting on auxiliary data, our
method (MTE + Full KLIEP) with word-vector embedding performs the best
overall -- including against 
\cite{Xu2015} which also exploits data augmentation; (iv) Finally, our
method is synergistic to the post processing
self-training approach \cite{Fu2015} as well as the hubness strategies
\cite{dinu2014improving}, which further explains the advantages of our
approach (MTE + Full KLIEP + PP) over other methods. 

\begin{table}[t]
\centering
\caption{Comparison versus state of the art. Embed: Label embedding, Feat: Visual feature used, Aug: Data augmentation required? TD: Transductive Requirement? }
\label{tab:ImportanceWeighting2}
\resizebox{0.72\linewidth}{!}{
\begin{tabular}{c|c|c|c|c|c|c|c}
\toprule
Method   & Embed & Feat & TD & Aug & HMDB51   & UCF101 & Olympic Sports                                         \\ \hline
MTE  & W & FV & \text{\sffamily X} & \text{\sffamily X}   &  $19.7\pm1.6$ & $15.8\pm1.3$ &  $44.3\pm8.1$ \\             
MTE + Full KLIEP & W & FV & \checkmark & \checkmark & ${23.9\pm 3.0}$ & ${21.9\pm 2.7}$ &  ${52.3\pm 8.1}$               \\
MTE + Full KLIEP + PP & W & FV & \checkmark & \checkmark & $\mathbf{24.8\pm 2.2}$ & $\mathbf{22.9\pm 3.3}$ & $\mathbf{56.6\pm 7.7}$             \\
MTE  & A & FV & \text{\sffamily X} & \text{\sffamily X}   &  N/A & $18.3\pm1.7$ &  $55.6\pm11.3$ \\             
\hline
DAP \cite{Lampert2009} - CVPR 2009  & A & FV & \text{\sffamily X} & \text{\sffamily X}   &       N/A        & $15.9\pm 1.2$ & $45.4\pm 12.8$           \\
IAP \cite{Lampert2009} - CVPR 2009 & A & FV & \text{\sffamily X} & \text{\sffamily X}    &      N/A        & $16.7\pm 1.1$ & $42.3\pm 12.5$           \\
HAA \cite{Liu2011}  - CVPR 2011 & A & FV &  \text{\sffamily X} & \text{\sffamily X}    &       N/A        & $14.9\pm 0.8$ & $46.1\pm 12.4$             \\ 
SVE  \cite{Xu2015} - ICIP 2015 & W & BoW & \text{\sffamily X} & \text{\sffamily X}  & $14.9\pm 1.8$ & $ 12.0\pm 1.4$ & N/A \\ 
SVE  \cite{Xu2015} - ICIP 2015 & W & BoW & \checkmark & \text{\sffamily X}  & $15.6\pm 0.7$ & $ 16.5\pm 2.4$ & N/A \\ 
SVE \cite{Xu2015} - ICIP 2015 & W & BoW & \text{\sffamily X} & \checkmark    & $19.3\pm 4.0$ & $13.1\pm 2.0$ & N/A                          \\
SVE \cite{Xu2015} - ICIP 2015 & W & BoW & \checkmark & \checkmark    & $22.8\pm 2.6$ & $18.4\pm 1.4$ & N/A                          \\

ESZSL \cite{Romera-paredes2015}  - ICML 2015 & W & FV & \text{\sffamily X} & \text{\sffamily X} &     $18.5\pm 2.0$        & $ 15.0\pm 1.3$  & $ 39.6 \pm 9.6$               \\
ESZSL \cite{Romera-paredes2015}  - ICML 2015 & W & FV & \text{\sffamily X} & \checkmark   &     $22.7\pm 3.5$        & $ 18.7\pm 1.6$  & $ 51.4 \pm 8.3$               \\

ESZSL \cite{Romera-paredes2015}  - ICML 2015 & A & FV & \text{\sffamily X} & \text{\sffamily X}   &     N/A        & $ 17.1\pm 1.2$  & $ 53.9 \pm 10.8$               \\
SJE \cite{Akata2015} - CVPR 2015 & W & FV & \text{\sffamily X} & \text{\sffamily X}   &     $13.3\pm 2.4$        & $ 9.9\pm 1.4$  & $ 28.6 \pm 4.9$               \\
SJE \cite{Akata2015} - CVPR 2015 & A & FV & \text{\sffamily X} & \text{\sffamily X}   &     N/A        & $ 12.0\pm 1.2$  & $ 47.5 \pm 14.8$               \\

UDA \cite{Kodirov2015} - ICCV 2015  & A & FV & \checkmark & \text{\sffamily X}        &     N/A        & $13.2\pm 1.9$  & N/A              \\ 
UDA \cite{Kodirov2015} - ICCV 2015  & A+W & FV & \checkmark & \text{\sffamily X}        &     N/A        & $14.0\pm 1.8$  & N/A              \\ 
\bottomrule
\end{tabular}}
\end{table}

\subsection{Qualitative Results and Further Analysis}

\begin{figure}[t]
\centering
{\includegraphics[width=.95\linewidth]{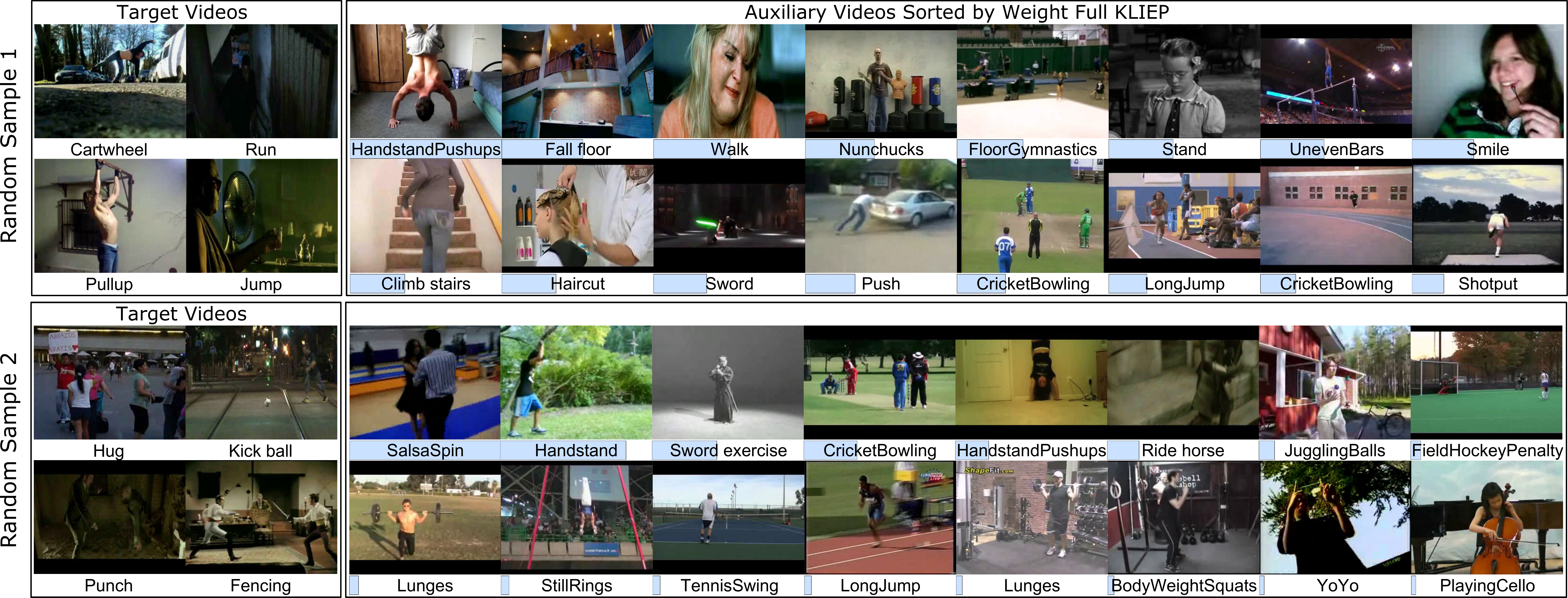}}
\caption{Visualisation of Full KLIEP auxiliary data weighting. Left: 4 target videos with category names. Right: 16 auxiliary videos with bars indicating the estimated weights.}\label{fig:QualitativeKLIEP}
\end{figure}

\noindent\textbf{Importance Weighting:}\quad To visualise the impact of our IW, we randomly select 4 / 16 classes as target / auxiliary sets respectively. We then estimate the weight on the 16 auxiliary video classes according to the Full KLIEP (Section~\ref{sect:ImportanceWeighting}). Examples of the auxiliary video weightings are presented in Fig~\ref{fig:QualitativeKLIEP}. We observe that auxiliary classes semantically related to the targets are given higher weight e.g. HandstandPushups$\to$Cartwheel in first sample, SalsaSpin$\to$Hug and Sword Exercise $\to$ Fencing in the second sample. While the visually and semantically less relevant auxiliary videos are given much lower weights.

\begin{figure}[t]
\centering
\includegraphics[width=0.88\linewidth]{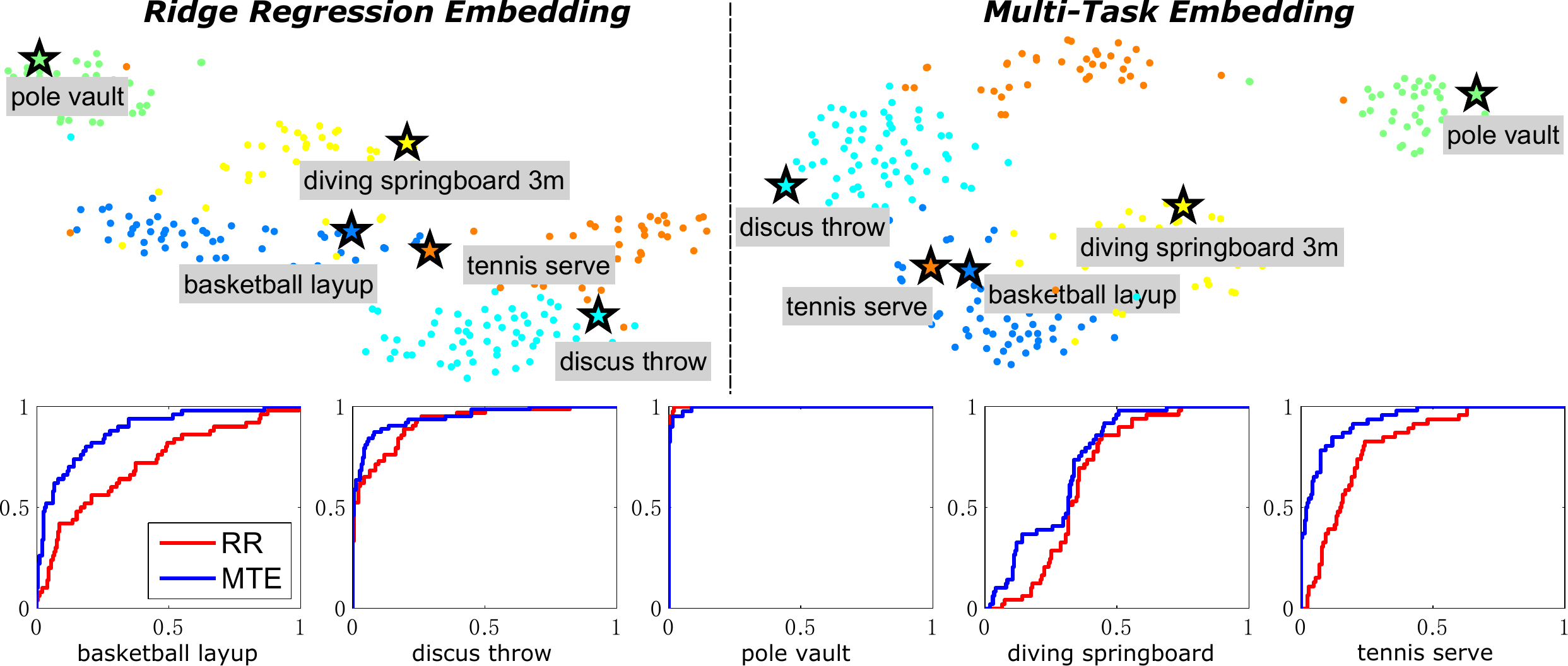}
\caption{Qualitative comparison between single-task ridge regression (RR) and multi-task embedding (MTE). 
}\label{fig:MTL_Qualitative}

\end{figure}

\noindent\textbf{Multi-task Embedding:}\quad We next qualitatively illustrate  single versus multi-task visual-semantic mappings. Specifically we take 5 classes to be
recognized and visualise their data after visual-semantic projection by
tSNE \cite{van2014accelerating}. A comparison between the 
representations generated by single-task (RR) and multi-task (MTE) 
mappings is given in Fig~\ref{fig:MTL_Qualitative}. 
The multi-task embedding discovers data in a lower dimension latent space where NN classification becomes more meaningful. 
The  improved representation is illustrated by computing the ROC curve for each target category, as seen in Fig~\ref{fig:MTL_Qualitative}. MTE provides improved detection over RR, demonstrating the better generalisation of this representation. 

\section{Conclusion}

In this work, we focused on zero-shot action recognition from the
perspective of improving generalisation of the 
visual-semantic mapping across the disjoint train/test class gap. We
propose both model- and data-centric improvements to a traditional
regression-based pipeline by respectively, multi-task embedding -- to
minimise overfit of the train data and to build a lower dimensional latent
matching space; and prioritising data augmentation by importance weighting -- to best
exploit auxiliary data for the recognition of target categories. Our
experiments on a set of contemporary action-recognition benchmarks
demonstrate the impact of both our contributions and show state-of-the-art results overall.

\bibliographystyle{splncs}
\bibliography{0342.bib}
\end{document}